**[Article Full Title]:** TransAnaNet: Transformer-based Anatomy Change Prediction Network for Head and Neck Cancer Patient Radiotherapy

**[Short Running Title]:** HNC Patient Anatomy Change Prediction

**[Author Names]:** Meixu Chen, PhD[1], Kai Wang, PhD[1, 2], Michael Dohopolski, MD[1], Howard Morgan, MD[1, 3], David Sher, MD[1], and Jing Wang, PhD[1]

**[Author Institutions]:**
[1] *Medical Artificial Intelligence and Automation (MAIA) Lab, Department of Radiation Oncology, UT Southwestern Medical Center, Dallas, TX, 75235, USA.*
[2] *Department of Radiation Oncology, University of Maryland Medical Center, Baltimore, MD, 21201, USA.*
[3] *Department of Radiation Oncology, Central Arkansas Radiation Therapy Institute, Little Rock, AR, 72205, USA.*

**[Corresponding Author Name & Email Address]:**
Corresponding Author: Jing Wang, PhD
Email: Jing.Wang@UTSouthwestern.edu
*Medical Artificial Intelligence and Automation (MAIA) Lab, Department of Radiation Oncology, UT Southwestern Medical Center, Dallas, TX, 75235, USA.*

**[Author Responsible for Statistical Analysis Name & Email Address]**
Meixu Chen, PhD
Email: Meixu.Chen@UTSouthwestern.edu

**[Conflict of Interest Statement for All Authors]**
*Conflict of Interest: None*

**[Funding Statement]**
*This work is supported by funding: NIH (R01 CA251792).*

**[Data Availability Statement for this Work]**
*Research data are not readily available because of institution regulations. Requests to access the datasets should be directed to corresponding author.*

**[Acknowledgements]**
We acknowledge the support from NIH (R01 CA251792).

# Abstract

**Background:** Adaptive radiotherapy (ART) can compensate for the dosimetric impact of anatomic change during radiotherapy of head neck cancer (HNC) patients. However, implementing ART universally poses challenges in clinical workflow and resource allocation, given the variability in patient response and the constraints of available resources. Therefore, early identification of head and neck cancer (HNC) patients who would experience significant anatomical change during radiotherapy (RT) is of importance to optimize patient clinical benefit and treatment resources.

**Purpose:** The purpose of this study is to assess the feasibility of using a vision-transformer (ViT) based neural network to predict radiotherapy induced anatomic change of HNC patients.

**Methods:** We retrospectively included 121 HNC patients treated with definitive RT/CRT. We collected the planning CT (pCT), planned dose, CBCTs acquired at the initial treatment (CBCT01) and fraction 21 (CBCT21), and primary tumor volume (GTVp) and involved nodal volume (GTVn) delineated on both pCT and CBCTs for model construction and evaluation. A UNet-style ViT network was designed to learn the spatial correspondence and contextual information from embedded image patches of CT, dose, CBCT01, GTVp, and GTVn. The deformation vector field between CBCT01 and CBCT21 was estimated by the model as the prediction of anatomic change, and deformed CBCT01 was used as the prediction of CBCT21. We also generated binary masks of GTVp, GTVn and patient body for volumetric change evaluation. We used data from 100 patients for training and validation, and the remaining 21 patients for testing. Image and volumetric similarity metrics including mean square error (MSE), structural similarity index (SSIM), dice coefficient, and average surface distance were used to measure the similarity between the target image and predicted CBCT.

**Results:** The predicted image from the proposed method yielded the best similarity to the real image (CBCT21) over pCT, CBCT01, and predicted CBCTs from other comparison models. The average MSE and SSIM between the normalized predicted CBCT to CBCT21 are 0.009 and 0.933, while the average dice coefficient between body mask, GTVp mask, and GTVn mask are 0.972, 0.792, and 0.821 respectively.

**Conclusions:** The proposed method showed promising performance for predicting radiotherapy induced anatomic change, which has the potential to assist in the decision making of HNC Adaptive RT.

# Introduction

Head and neck cancer (HNC) is one of the most common types of cancer in the United States, with an estimated 66,920 new cases anticipated in 2023.[1,2] In current clinical practice, the predominant approach to managing locally advanced HNC involves the application of radiotherapy (RT), with intensity modulated radiotherapy (IMRT) serving as the standard treatment technique because of its highly conformal dose distribution and steep dose gradient. Nevertheless, a subset of HNC patients experience substantial anatomical and geometric variations in both target volumes and organs at risk (OARs) during IMRT. These deviations may lead to inadvertent overdosing of normal tissue and/or underdosing of target structures, thereby compromising the therapeutic benefits of this highly conformal approach. Previous studies have suggested that the affected HNC patient population can range from 21% to 66%.[3-6] While image-guided radiotherapy (IGRT) is routinely employed in most treatment centers to correct day-to-day positional deviations between planning computed tomography (CT) and daily images, it remains inadequate for addressing internal anatomical changes of target or normal tissues.

Adaptive radiotherapy (ART) is desired for mitigating unfavorable dosimetry outcomes for HNC patients experiencing anatomical changes during IMRT. The dosimetric benefits and favorable clinical outcomes stemming from ART have been extensively reported.[3,5,7,8] However, the extensive adoption of ART for all patients is challenging. As a resource intensive technique, it requires time-consuming and labor-intensive procedures including repeated imaging, contouring, replanning, and dosimetric analysis.[3,6,9,10] Therefore, accurately identifying individuals who would benefit from adaptive replanning or receiving online ART treatment based on patient-specific dosimetric advantages is crucial for effective clinical resource management.

Numerous investigations have been conducted to explore various clinical and imaging factors that predict the necessity for replanning in ART. These studies have aimed to establish criteria for identifying anatomical changes that could result in suboptimal dosimetric outcomes HNC RT. Statistical analyses have pinpointed several potential indicators, including tumor location, patient age, Body Mass Index (BMI), intended dosage to the parotid glands, initial volume of the parotid glands, and the extent of their overlap with the planning target volume (PTV).[3,4,7,11,12] More recent advances in radiomics and machine learning, have led to the development of multi-variable binary prediction models specifically for selecting HNC ART patients.[13-16] These models, trained using pre-treatment CT or MRI radiomics data, with or without additional clinical factors, have demonstrated superior binary prediction accuracy compared to models based solely on clinical factors. However, the absence of a standardized clinical protocol for determining ART eligibility poses significant challenges for the cross-institutional application of these models. Further, the limited interpretability of radiomic models and binary prediction outcomes potentially diminishes their reliability for clinical application. Consequently, these models do not provide insights into the primary causes of dose variation nor guidance for clinicians during the replanning process.

Different from binary prediction, image prediction of anatomy change could offer a more substantial and perceptible set of prognostic data for physicians in the decision making of ART. Deep learning-based image prediction and generation has recently emerged as a focal area of interest, spurred by the remarkable advancements in Generative Adversarial Networks (GAN) and Vision Transformer-based networks (ViT). This technological evolution has culminated in the development of a spectrum of research models and even commercial products dedicated to tasks such as text-image translation, image inpainting, motion prediction, and next video frame prediction.[17-21] In the field of oncology, inspired by the achievement in general image prediction tasks, several studies have demonstrated promising image-based tumor anatomy change prediction results on different cancer sites, including brain, lung, and pancreas, for either tumor growth prediction or tumor treatment response prediction.[22-25] Their prediction results presenting in the form of 3D images show enhanced interpretability and can be used for image segmentation, tumor aggressiveness quantification, treatment response prediction, or assisting in determining the necessity for RT replanning.

In this study, we hypothesize that a ViT-based deep image prediction model (TransAnaNet) constructed by multifaceted analyses of planning CT, daily Cone Beam CT (CBCT) and dosimetric data of HNC patients is indicative of patient anatomy change during RT, which can assist in the decision making of HNC ART. Previous studies indicated that the optimal timing for replanning is around the third and the fourth week of treatment on the off-line adaptive setting.[26-28] Based on these findings, we selected the patient anatomy change at the end of the fourth week (fraction 21th) as the prediction target of this study. We leveraged a retrospective database to investigate the accuracy and characteristics of the anatomy change prediction model. To the best of our knowledge, this is the first deep learning model for predicting HNC patient anatomy change during RT.

## Methods

### A. Dataset and preprocessing

#### *Patient and Data*

This retrospective study was reviewed and approved through the institutional review board (IRB). Patients treated between April 2014 and October 2019 at the University of Texas Medical Center were included if they were diagnosed with locally advanced HNC (including oropharynx, supraglottic, glottic, or hypopharynx) and completed a full course of conventionally fractionated definitive radiotherapy with daily or weekly CBCT imaging. Patients with all primary structures and nodal structures receiving 70 Gy dose were selected for evaluation. Those who had a prior history of RT to the head and neck, received prior induction chemotherapy, had distant metastases (DM), or had the presence of a separate active malignancy, were excluded.

We collected 4 sets of image data for each of the patients for analysis: the baseline CT simulation scan (CT), the 3D dose map from the treatment planning system, CBCT prior to the initial fraction (CBCT01), and CBCT prior to fraction 21 (CBCT21). At baseline simulation, all patients were simulated on a Philips 16-slice Brilliance large-bore CT simulator with iodinated IV contrast. CBCT imaging was performed either daily or weekly on a Varian TrueBeam™ machine (Varian Medical Systems, Palo Alto, CA). See supplementary file for more details about imaging parameters (Table S1).

### *Data preprocessing.*

All initial segmentations, including patient body contours, on CT were delineated by a board-certified radiation oncologist specializing in head and neck radiotherapy. Nodal contours were combined into a single structure to simplify the analysis. For CBCT images, all segmentations were deformed from CT with rigid and/or deformable image registration in Velocity (Varian Medical Systems, Palo Alto, CA). All generated contours were manually edited by the physician to verify the inclusion of all affected mucosa if applicable and to exclude any incident bone, air, cartilage, or adipose tissue that may be overlapping the GTV boundaries. This process was repeated for both CBCT01 and CBCT21. Then, all the image data including dose map and segmented RT structure masks were rigid registered to CBCT21, the voxel sizes were resampled to $2\times2\times2$ mm$^3$, and the images were center cropped to $128\times128\times32$ matrixes for the model training process. CT and CBCT images were clipped to [-1000, 1000] HU and min-max normalized to [-1, 1], dose maps were z-score normalized, segmentation masks were binarized to 0 and 1.

## B. Transformer-based Anatomy Prediction Network

### *Workflow*

Recently, vision transformer (ViT) architectures have been applied to medical image registration tasks. Comparing with convolutional neural networks (CNNs), ViT models could predict more precise deformation vector fields (DVFs) between the moving and fixed images due to the capability of capturing long-range spatial information.[29-31] Inspired by the success of ViT-based method for medical image registration, we expanded this idea from spatial relationship prediction between two known images (image registration), to spatial-temporal relationship prediction between known images and future images (anatomy change prediction).

The proposed framework of using transformer for patient anatomy change prediction is shown in Figure 1(a). A ViT-encode and a convolution decoder formed the hybrid Transformer-ConvVNet (TransAnaNet) for anatomy change prediction. The hybrid network takes seven inputs: planning CT image, GTVp mask on planning CT, GTVn mask on planning CT, planned dose map, initial fraction CBCT (CBCT01), and GTVp and GTVn mask on it. The network predicts a nonlinear warping DVF as the intermediate output, which is then applied to CBCT01 image through a spatial transformation function to generate the predicted patient anatomy (CBCT 21). The GTVp and GTVn masks on CBCT21 are also predicted by deforming the corresponding masks on CBCT01. During model training stage, the image similarity and structure similarity between the predicted and real CBCT21 are used as part of the loss functions to optimize the model, while in the model validation stage they are also used to evaluate the performance of the constructed model. Of note, here we set CBCT01 as the baseline image for deformation since it shares the same imaging protocol as the prediction target CBCT21. It can be replaced with planning CT or any other early fraction CBCTs for anatomy change prediction as long as the dataset is consistent. On the other hand, the target image can be replaced by other late fraction CBCTs according to the needed prediction time point.

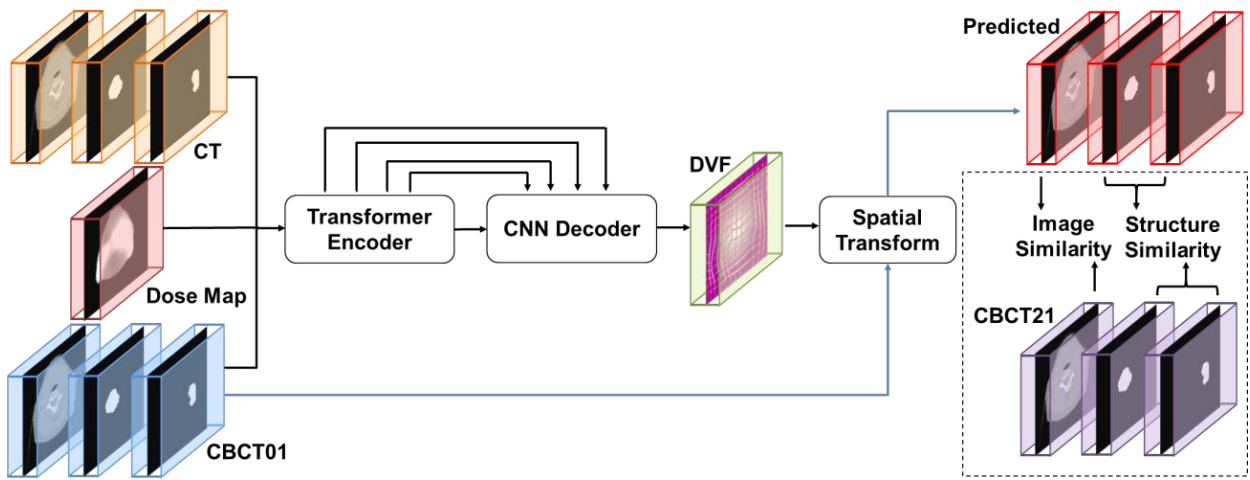

(a)

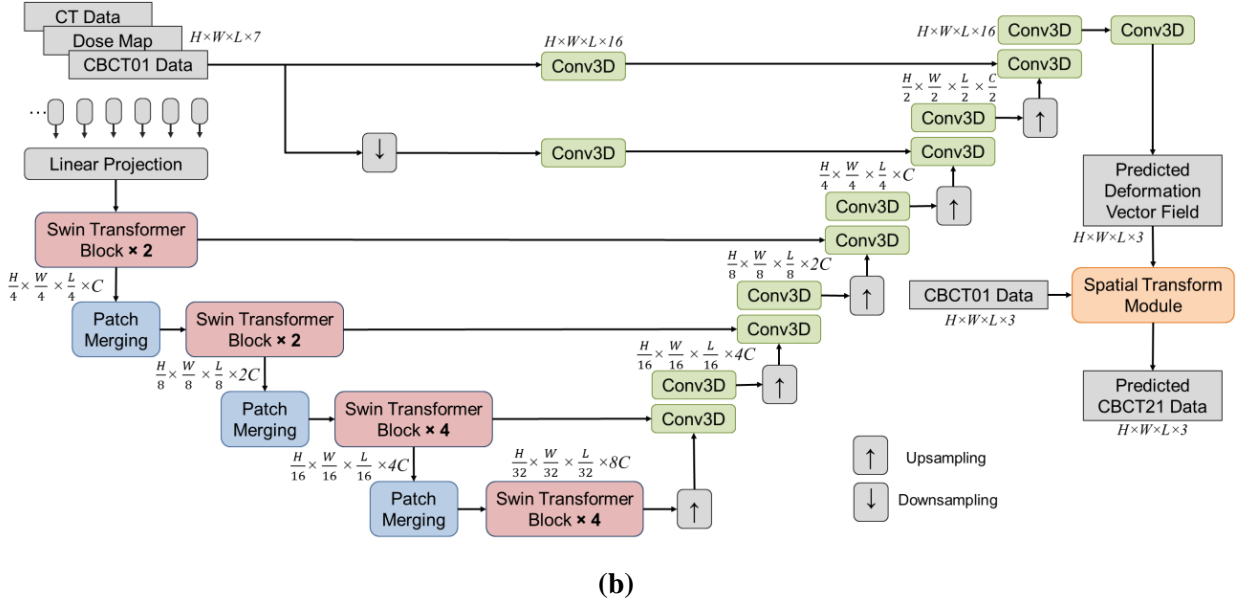

**(b)**

**Figure 1.** The overall framework (a) and detailed architecture (b) of the proposed Transformer-based head and neck cancer patient anatomy change prediction (TransAnaNet) model. The hybrid Transformer-ConvNet network takes seven inputs: planning CT image, GTVp mask on planning CT, GTVn mask on planning CT, planned dose map, initial fraction CBCT (CBCT01), and GTVp and GTVn mask on it. The network predicts a nonlinear warping deformation vector field (DVF), which is then applied to CBCT01 image through a spatial transformation function to generate the predicted patient anatomy (fraction 21). For training data, the image similarity and structure similarity between the predicted and real CBCT21 are used as part of the loss functions to update the model, they are also used to evaluate the performance of the constructed model.

### Network Structure

Figure 1(b) shows the network architecture of the proposed TransAnaNet. The model structure was modified based on a deep medical image registration model named TransMorph, which has state-of-the-art performance on image registration tasks.[31] We used the Swin Transformer as ViT encoder to capture the spatial correlation between different regions in the input dose map, planning CT, CBCT01 images, and their GTV masks. Comparing to previous version of ViT module, Swin Transformer module can build hierarchical feature maps by merging image patches in deeper layers and has linear computation complexity to input image size, which is of high efficiency for model training.[32,33] Then a typical convolutional decoder is used to process the information provided from the transformer encoder into a dense displacement hidden field. Cross-layer connections were used to maintain the localization information between the encoder and decoder.

The encoder of the network first splits the input image, dose, and mask volumes into non-overlapping 3D patches, each of size $7 \times P \times P \times P$. Then each patch is flattened and linear projected to a feature representation of dimension $L{\times}C$ via the linear projector, where $L{=}\frac{H}{P}\times\frac{W}{P}\times\frac{L}{P}$ is the total number of patches ($H{\times}W{\times}L{\times}7{\rightarrow}\frac{H}{4}\times\frac{W}{4}\times\frac{L}{4}\times C$, we set $P$ to 4 following the typical setting). Following the linear projection, several consecutive modules of patch merging and Swin Transformer blocks are adopted. The Swin Transformer blocks outputs the same number of features as the input, while the patch merging layers merge the features of each group of 2×2×2 neighbors, thus they reduce the number of tokens by a factor of 2×2×2=8. Then a linear layer is applied to produce features each of $2C$-dimension ($\frac{H}{4}\times\frac{W}{4}\times\frac{L}{4}\times C{\rightarrow}\frac{H}{8}\times\frac{W}{8}\times\frac{L}{8}\times 2C$). After four layers of Swin Transformer blocks and three patch merging modules in between, the features are encoder to dimension of $\frac{H}{32}\times\frac{W}{32}\times\frac{L}{32}\times 8C$. The decoder consists of multiple upsampling and convolutional layers with the kernel size of $3\times3$. Each of the upsampled feature maps in the decoding stage was concatenated with the corresponding feature map from the encoding path via skip connections, then two convolutional layers were applied. We also employed two convolutional layers using the original and downsampled image pair as inputs to capture local information and generate high-resolution feature maps. The outputs of these layers were concatenated with the feature maps in the decoder to produce a DVF. Leaky Rectified Linear Unit is adopted following each convolution layer except for the last DVF generation convolutional layer. Finally, a spatial transform module takes the predicted DVF and CBCT01 data as input and generate the predicted CBCT21 Data, which comprises of the predicted CBCT21 image, the predicted GTVp mask, and GTVn mask. During model training, the spatial transformation module apply a differentiable nonlinear warp using the predicted DVF to the deform input images (CBCT01 data), while it uses tri-linear interpolation to deform images (CBCT21) and nearest-neighbor interpolation to deform mask images (GTVp and GTVn) during inference.[34]

### *Loss Functions*

The loss function we used for model training is combined with similarity loss $L_{similarity}$ and diffusion loss $L_{diffusion}$:

$$Loss = L_{similarity}([\mathcal{D}(I_{01}, \phi), I_{21}], [\mathcal{D}(P_{01}, \phi), P_{21}], [\mathcal{D}(N_{01}, \phi), N_{21}]) + \lambda L_{diffusion}(\phi)$$

where $I_n$ is the CBCT image collected at fraction #$n$, $P_n$ and $N_n$ are the corresponding GTVp volume and GTVn volume, $\phi$ is the estimated DVF, $\mathcal{D}(x, y)$ denotes the deformed moving image $x$ with DVF $y$. The similarity loss consists of overall image similarity loss between predicted image and target image, and structural similarity loss between the predicted GTV and GTV of target image. We used SSIM loss as the similarity loss for CBCT images, and Dice coefficient loss as the structural similarity loss of GTVp and GTVn. We assigned weighting factor of value 1.0 to each of the similarity loss empirically. DVF gradient was used as the diffusion loss in our study, $\lambda$ was chose as 0.01 to balance the accuracy and smoothness of the predicted deformation field.

### *Model Performance Evaluation*

We used a testing dataset for model performance evaluation. To demonstrate the effectiveness of the proposed model for anatomy change prediction, we compared the similarity between the target image (CBCT21) and predicted image to the similarity between CBCT21 and other images, including planning CT, CBCT01. Two other deep prediction models using different encoders were also trained for comparison. One of them replaced our Swin Transformer encoder with Convolutional Neural Network (CNN) to mimic the structure of VoxelMorph, which is a widely used CNN-based image registration network.[35] The other model replaced the Swin Transformer block with basic ViT block[32] The comparison was conducted quantitatively and qualitatively.

For quantitative evaluation, consistent with the design of our loss function, we evaluated the accuracy of anatomy prediction in two aspects, overall image similarity and structural similarity to CBCT21. Mean square error (MSE) and structure similarity index (SSIM) were applied to the whole 3D image to quantify the similarity of overall image. Dice coefficient (Dice) and average symmetric surface distance (ASD) were applied to patient body mask, GTVp and GTVn masks to quantify the structural similarity. We averaged the scores of all testing patients for comparison. The mean and standard deviation of each metrics were compared across CT, CBCT01 and predicted images.

For qualitative evaluation, we visually checked the similarity and image/structure deformation between CT, CBCT01, and predicted CBCT21 to real CBCT21. An accurate anatomy change prediction is expected to generate an image which looks more similar to CBCT21 anatomically but not CT or CBCT01, and it could reflect where significant anatomy change occurs. We showed the CT, CBCT01, CBCT21, and predicted CBCT21 to demonstrate that. The difference images between the body masks of different images to that of CBCT21 were also listed for inspection.

A set of ablation studies were done to evaluate the contribution of each input image modality. We removed CT, Dose, and GTV masks from the input data in succession and retained our model accordingly. The mean and standard deviation of each metric for them were reported and compared. In addition, we changed the baseline image from CBCT01 to CT to investigate the impact of the choice of baseline image on the prediction results.

*Implementation Details*

Training and validation code of the proposed method were implemented using PyTorch on a workstation with two NVIDIA RTX3090 GPUs. We trained all the models for 100 epochs using the Adam optimization algorithm with ReduceLROnPlateau strategy, the initial learning rate is set as 0.001 and the training batch size is 4. The input images were augmented with flipping, shifting, rotating and added Gaussian noise. Swin transformer patch size was set to [2, 4, 4], window size was set as [5, 5, 5], block numbers were set as [2, 2, 4, 2], head number as [4, 4, 8, 8], and embedding dimension as 96. For the other two comparison deep learning model, we followed the default settings of model hyper parameters in their published implementations.[32,35]

# Results

## A. Patient Data

One-hundred twenty-one eligible patients were included in the current study, their demographic, disease characteristics, and treatment related information are summarized in Table 1. Examples of their planning CT, CBCT01, and CBCT21 are provided in the supplementary file. Distribution of patient BMI and change of BMI are also presented in the supplementary. Of note, during review of imaging, the image quality of the GTVp and/or GTVn on some CBCTs were found to be degraded by strong artifacts. Therefore, patients with CBCT01 or CBCT21 scans affected were replaced with a separate repeat scan done prior to the same fraction, preferably, or a CBCT +/− 1 fraction if a repeat scan was not available. In the presenting study, patients were randomly stratified into training (n = 80), validation (n = 20) and testing (n = 21) sets for model training and evaluation.

**Table 1.** Patient and treatment characteristics.

| Characteristics | | Total=121 |
|---|---|---|
| | | Number / Median [IQR] |
| Gender | Male | 100 |
| | Female | 21 |
| Ethnicity | White | 77 |
| | African American | 19 |
| | Hispanic | 9 |
| | Asian | 2 |
| | Other and Unknown | 14 |
| Age | | 61 [55~67] |
| BMI | | 28.3 [25.3~31.2] |
| Disease Site | Oropharynx | 68 |
| | larynx | 37 |
| | Oral cavity | 2 |
| | Other | 14 |
| Disease Laterality | Left | 35 |
| | Right | 35 |
| | Central | 22 |
| | Bilateral | 29 |
| T Stage | 1 | 18 |
| | 2 | 28 |
| | 3 | 45 |
| | 4 | 20 |
| N Stage | 0 | 20 |
| | 1 | 31 |
| | 2 | 66 |
| | 3 | 4 |
| HPV-P16 Status | Positive | 60 |
| | Negative | 21 |
| | Unknown | 40 |
| Smoking Status | Never smoker | 37 |
| | Former smoker | 54 |
| | Current smoker | 21 |
| | Unknown | 9 |
| Treatment Paradigm | Chemoradiotherapy | 114 |
| | Radiotherapy | 7 |
| RT Modality | IMRT | 121 |
| Dose and Fraction | 7000 cGy in 35 Fx | 75 |
| | 6996 cGy in 33 Fx | 46 |
| Days between CT-Sim to RT | | 12 [7, 18] |
| Days between Fx#1 to Fx#21 | | 28 [27, 30] |

## B. Validation of TransAnaNet Model

Table 2 summarizes the image and structure similarity between CBCT21 to planning CT, CBCT01, and predicted CBCT21 from different anatomy change prediction models. The mean value and standard deviation of each evaluation metric on the testing cohort were reported. Example images and GTV contours of three sample testing patients (patients with the smallest, median, and the largest volumetric change between planning CT and CBCT21) are shown in Figure 2. Difference images between body masks of each image to those of CBCT21 were also measured to show the anatomy change. Distribution of the evaluation metrics on testing cohort are summarized in the supplementary.

According to Table 2, the predicted CBCT21 images from the proposed TransAnaNet have the best overall image similarity to real CBCT21 scans in terms of MSE ($0.009\pm0.003$), and SSIM ($0.933\pm0.020$). For structure similarity, the results from the proposed TransAnaNet still outperform other scans or predicted images in body contour (Dice = $0.972\pm0.008$, ASD = $1.910\pm0.318$), GTVp contour (Dice = $0.792\pm0.090$, ASD = $1.338\pm0.398$), and GTVn contour (Dice = $0.821\pm0.130$, ASD = $0.922\pm0.720$). The predicted images from CNN have the same performance on MSE image similarity as TransAnaNet, but lower performance on other metrics.

Different degrees of anatomy change can be identified from Figure 2 when comparing CT/CBCT01 to CBCT21 or checking the difference body mask images between CBCT21 to CT/CBCT01. Comparing with CT, the predicted CBCT21 images from the proposed TansAnaNet have similar image visual quality to CBCT21 as they were deformed from CBCT01, which were collected with the same machine and protocol as CBCT21. Comparing with both CT and CBCT01, the predicted CBCT21 images have less body volume difference to CBCT21.

**Table 2.** Image and structure similarity between CBCT21 to planning CT, CBCT01, and predicted CBCT21 from different anatomy change prediction models for the testing cohort (21 patients, median values along with their corresponding standard deviations are reported). (a) The image similarity is quantified with mean square error (MSE) and structure similarity index (SSIM). (b) The structure similarity is quantified with dice coefficient (dice) and average symmetric surface distance (ASD, unit: mm) for binary masks of patients' head and neck regions (body), primary tumors (GTVp) and involved lymph nodes (GTVn). Anatomy change prediction models built with Swin Transformer-based encoder (TransAnaNet), convolutional neural network based encoder (CNN), and basic vision-transformer-based encoder (ViT) are listed for comparison. Best performance for each metric is bolded.

(a)

| Image/Model | Image Similarity | |
| --- | --- | --- |
| | MSE | SSIM |
| Planning CT | 0.016±0.007 | 0.877±0.048 |
| CBCT01 | 0.013±0.004 | 0.919±0.028 |
| TransAnaNet | **0.009±0.003** | **0.933±0.020** |
| CNN | **0.009±0.003** | 0.932±0.021 |
| ViT | 0.010±0.003 | 0.930±0.020 |

(b)

| Image/Model | Structure Similarity | | | | | |
| --- | --- | --- | --- | --- | --- | --- |
| | Body | | GTVp | | GTVn | |
| | Dice | ASD | Dice | ASD | Dice | ASD |
| Planning CT | 0.948±0.019 | 2.914±1.062 | 0.731±0.083 | 1.480±0.390 | 0.760±0.164 | 1.194±0.880 |
| CBCT01 | 0.961±0.014 | 2.210±0.684 | 0.741±0.099 | 1.462±0.562 | 0.786±0.159 | 1.084±0.836 |
| TransAnaNet | **0.972±0.008** | **1.910±0.318** | **0.792±0.090** | **1.338±0.398** | **0.821±0.130** | **0.922±0.720** |
| CNN | 0.968±0.009 | 1.946±0.328 | 0.778±0.088 | 1.402±0.454 | 0.814±0.134 | 0.984±0.768 |
| ViT | 0.969±0.008 | 1.924±0.294 | 0.781±0.094 | 1.338±0.474 | 0.815±0.143 | 0.974±0.782 |

**Figure 2.** Qualitative performance evaluation of the proposed method for anatomy change prediction (4 example patients out of the 21 testing cohort patients). Testing patients with a) the smallest, b) median, and c) the largest volumetric change between planning CT and CBCT21 are selected for comparison. CT, CBCT01, CBCT21 and predicted CBCT21 images are listed for comparison, original and predicted GTVp and GTVn are delineated in red and blue respectively. Difference images between the body masks of planning CT, CBCT01, and predicted CBCT21 to those from the real CBCT21 are shown in the second row for each patient.

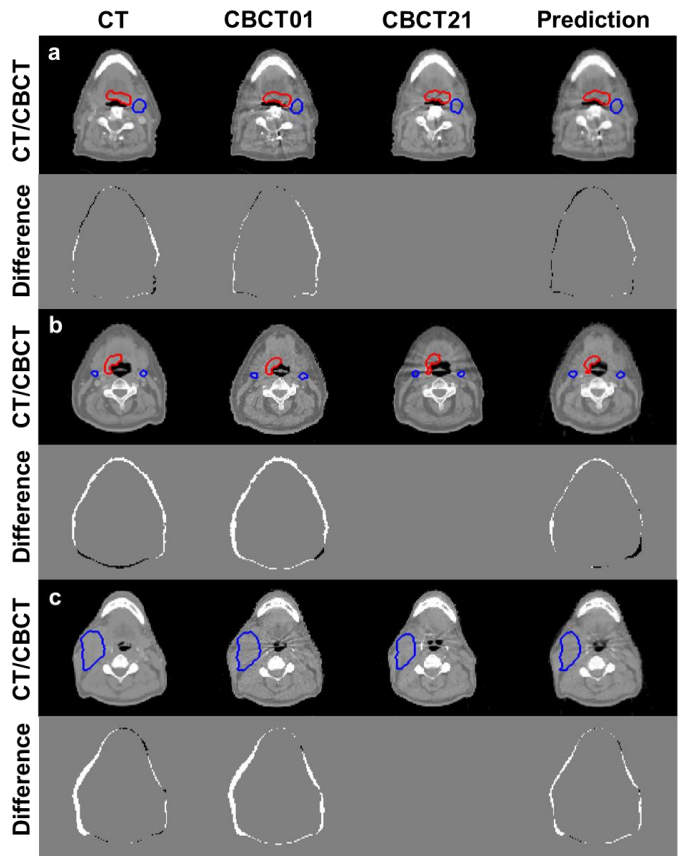

## C. Ablation Study

Table 3 compares the performance of anatomy change prediction using different baseline images and combination of image modalities as input for the proposed TransAnaNet model. The result shows that all of the collected image modalities have positive contribution to the anatomy change prediction using the proposed model. Missing either CT image, GTV mask, planned dose, or CBCT01 as input data will lead to degraded performance. When adopting CBCT01 as the baseline image for deformation and using the combination of 3D dose map, planning CT, CBCT01, and GTV contours on CT and CBCT01 as model input, the proposed network has the best performance on overall image similarity (MSE: 0.009±0.003, SSIM: 0.933±0.020), and it has the superior head and neck body mask structure similarity to CBCT21.

The best GTV structure similarity performance was achieved when using the same input image to the model, but adopting planning CT as the baseline image for deformation. The resulting GTVp similarity performance is 0.807±0.084 and 1.296±0.364 for dice coefficient and ASD respectively. The resulting GTVn similarity performance is 0.829±0.128 and 0.904±0.688 for dice coefficient and ASD respectively. Some testing samples generated following this input modality combination and baseline image are shown in Figure 3. The generated CBCT21 images have similar widths of body contours as CBCT21, and some of the volumetric changes of sub-structures can be clearly identified from the images (pointed by red arrows in Figure 3).

**Table 3.** Performance comparison of using different baseline image and combination of image modalities as input for the proposed anatomy prediction model on the testing cohort (21 patients, median values along with their corresponding standard deviations are reported). a) The image similarity is quantified with mean square error (MSE) and structure similarity index (SSIM). b) The structure similarity is quantified with dice coefficient (dice) and average symmetric surface distance (ASD, unit: mm) for binary masks of patients' head and neck regions (body), primary tumors (GTVp) and involved lymph nodes (GTVn). Best performance for each metric is bolded.

(a)

| Image Modality | Image Similarity | |
|---|---|---|
| | MSE | SSIM |
| CBCT01+GTV+Dose (Baseline: CBCT01) | 0.011±0.004 | 0.925±0.027 |
| CT+CBCT01+GTV (Baseline: CBCT01) | 0.010±0.003 | 0.929±0.022 |
| CT+CBCT01+Dose (Baseline: CBCT01) | 0.010±0.003 | 0.929±0.022 |
| CT+CBCT01+GTV+Dose (Baseline: CT) | 0.014±0.004 | 0.902±0.035 |
| CT+CBCT01+GTV+Dose (Baseline: CBCT01) | **0.009±0.003** | **0.933±0.020** |

(b)

| Image/Model | Structure Similarity | | | | | |
|---|---|---|---|---|---|---|
| | Body | | GTVp | | GTVn | |
| | Dice | ASD | Dice | ASD | Dice | ASD |
| CBCT01+GTV+Dose (Baseline: CBCT01) | 0.960±0.017 | 2.187±0.697 | 0.749±0.091 | 1.436±0.404 | 0.787±0.160 | 1.050±0.802 |
| CT+CBCT01+GTV (Baseline: CBCT01) | 0.971±0.010 | 2.004±0.357 | 0.775±0.087 | 1.422±0.430 | 0.805±0.145 | 0.992±0.779 |
| CT+CBCT01+Dose (Baseline: CBCT01) | 0.970±0.010 | 2.024±0.346 | 0.744±0.091 | 1.486±0.512 | 0.782±0.155 | 1.092±0.816 |
| CT+CBCT01+GTV+Dose (Baseline: CT) | 0.971±0.009 | 1.954±0.342 | **0.807±0.084** | **1.296±0.364** | **0.829±0.128** | **0.904±0.688** |
| CT+CBCT01+GTV+Dose (Baseline: CBCT01) | **0.972±0.008** | **1.910±0.318** | 0.792±0.090 | 1.338±0.398 | 0.821±0.130 | 0.922±0.720 |

**Figure 3.** Qualitative performance evaluation of the proposed method for anatomy change prediction with the planning CT image as the baseline image. Planning CT, CBCT01, CBCT21 and predicted CBCT21 from the proposed method for seven different testing patients (columns a~g) are listed as examples. The dash lines aside of each set of images show the width of the corresponding CBCT21 body contours, the red arrows point to the regions of obvious volumetric changes.

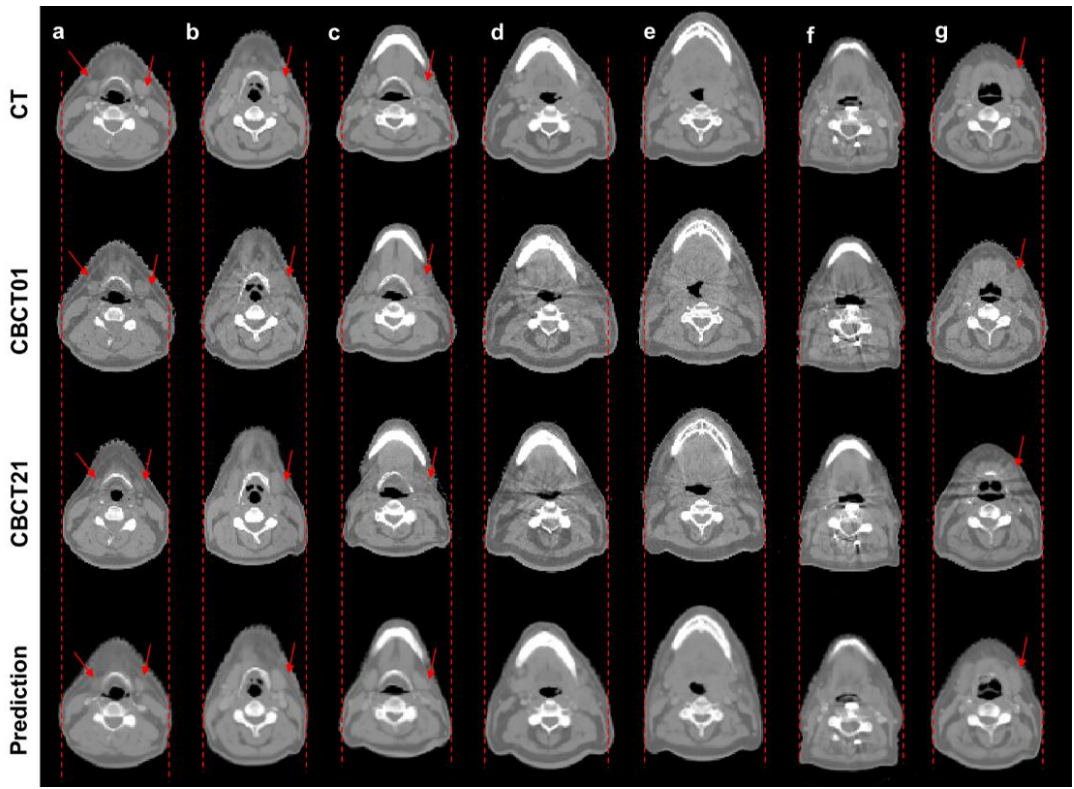

## Discussion

During HNC IMRT courses, patients may undergo substantial anatomical changes, resulting in a divergence of the actual administered dose from the initial treatment plan during the later stages of therapy, which might lead to unexpected side effects or loss of tumor control. In this exploratory study, we proposed the first deep learning model prognosticate patient anatomy change in HNC radiotherapy, thereby facilitating the early detection of patients susceptible to significant anatomical changes. We demonstrated that a ViT model (TransAnaNet) with planned dose, planning CT, early fraction CBCT, and GTV masks as input possessed high predictive ability for patient anatomy change in later treatment fraction. The proposed method shows great potential in assisting clinicians to identify HNC patients who are likely to derive greater benefit from ART. The code to implement our method is publicly available on GitHub (Link will be released after acceptance).

We propose to use initial CBCT (CBCT01) as the baseline image, and predict the DVF between initial CBCT and late fraction CBCT (CBCT21). Subsequently, CBCT01 is deformed according to the predicted DVF to generate the predicted CBCT21. Comparative analysis with the planning CT and CBCT01 indicates that the predicted CBCT21 more closely approximates the actual CBCT21 in terms of overall image similarity and structural similarity (Table 2). The MSE and SSIM between the predicted CBCT21 to real

CBCT21 are 0.009 and 0.933 respectively. The dice coefficient and ASD (unit: mm) for body mask, GTVp mask, and GTVn mask are 0.972/1.910, 0.792/1.338, and 0.821/0.922 respectively. We also compared the performance derived from other deep learning models, and the proposed Swin Transformer based model has the best performance for all metrics.

A set of ablation studies were done to evaluate the contributory significance of each input image modality (Table 3). The result shows that every image modality employed enhanced the anatomy change prediction accuracy of the proposed model. Omitting any single modality results in diminished performance. While the overall image similarity and body mask similarity are not sensitive to input modalities, the GTVp and GTVn mask similarity are sensitive to input image data. Notably, the most significant decrease in performance is observed when the input lacks the preceding GTV mask. In this case, the predicted DVF only deforms the boundary of patient body, without predicting the anatomy change of internal sub-structures. This outcome underscores the essentiality of incorporating sub-structure similarity within the loss function to refine model optimization.

Of note, due to the heavy workload of contouring OARs on CBCTs, in current study, we constructed our loss function and evaluated our model based on the head and neck 3D image and GTV masks. However, existing literature suggests that anatomical changes result in greater dose variation in OARs as opposed to target volumes.[4,6,12] The dose coverage of the GTV is typically more resilient to these changes due to the incorporation of the PTV concept. On the other hand, the planning volumes at risk (PRV) margins are only commonly employed for the spinal cord and brain stem, not for most of the other main OARs in HNC RT, such as the parotid glands which are consistently reported to shrink and/or appear orientation shift during treatment.[3,4,36,37] Therefore, the incorporation of OAR contours and the refinement of our model to include these structures could enhance the efficacy of the proposed methodology, which is worthy of exploring in a future study.

In our experiment, we also validate the performance of the proposed model when planning CT is used as the baseline image (Table 3, Figure 3), where the model predicts the DVF between planning CT and late fraction CBCT. It turns out that using planning CT as the baseline image achieved the best sub-prediction accuracy. The resulting dice coefficient and ASD (unit: mm) is 0.807/1.296 and 0.829/0.904 for GTVp and GTVn respectively. Concurrently, using CBCT01 as the baseline has better performance in terms of overall image and body mask similarity. The presence of artifacts in CBCT and the more pronounced soft tissue contrast in CT may account for the enhanced performance in sub-structure anatomical prediction when CT is utilized as the baseline image. Additionally, the consistency of imaging protocols could explain the heightened overall image similarity when CBCT01 was used as the baseline. Selection of baseline image merits additional investigation.

Previous studies have introduced various machine learning and radiomics techniques for predicting ART eligibility of HNC patients with the hypothesis that features from multiple modalities contain predictive biomarkers for tumor and OAR shrinkage following cancer treatment.[13-16] However, despite their high accuracy in binary prediction, these methods face challenges in quantifying the anatomical changes of tumors and/or OARs. Furthermore, given the variability in clinical guidelines for ART across different institutions, the blackbox nature of these methods may undermine their reliability and reproducibility when applied across diverse clinical settings, and the straightforward binary prediction yields low interpretability.[38-40] In contrast, we aimed to build an anatomy change prediction model that does not directly predict the necessity for ART but provides the possible anatomy change of the treatment volume through image data, thereby assisting the decision making for ART.

The proposed method facilitates visualization of potential anatomical changes, which is a pivotal advantage over previous methods. Through our method, the predicted patient anatomy change can be visualized via the predicted DVFs, the predicted images, or the structure mask difference maps (Figures 2 and 3). Independent of whether CBCT01 or CT serves as the baseline image, the predicted image can distinctly delineate the location and extent of anatomy changes. This demonstrated the effectiveness of the proposed anatomy change prediction method qualitatively.

The application of the proposed model extends beyond predicting later fraction CBCT images. Its primary function is to identify optimal candidates for ART by evaluating the magnitude of anatomical changes. In addition, with an accurate prediction of patient anatomy change, the current treatment plan can be dosimetrically evaluated on the predicted patient anatomy, thereby informing physicians' decisions regarding the necessity for re-simulation or re-planning to align with clinical objectives. On the other hand, for online ART, the predicted image could potentially streamline the preparation of new plan for next treatment, which might reduce the duration of patient wait times on the treatment couch and enhance the efficiency of the clinical workflow.

As a proof-of-concept study, our work has several limitations. First, the anatomy changes were predicted at a fixed fraction (fraction 21), which might not be the optimal time point decision making or replanning. Constructing a longitudinal prediction model using recurrent neural network to predict the anatomy change in time-series will be one of the main directions of our future work. Additionally, the model's construction and validation were based on a limited patient cohort, and there is currently no independent data from an external institution or collected prospectively for evaluation. Conducting prospective study for model evaluation is envisaged for our future work, and the code for implementation of the proposed method is publicly available for external use. Thirdly, the target delineation quality, interobserver variability and their potential influence were not evaluated. Finally, although anatomy change prediction offers physicians more detailed information to assist in the decision making of replanning or ART than binary prediction methods, the prediction uncertainty of the proposed TransAnaNet has yet to be scrutinized. Our future research efforts may also involve exploring the model's interpretability and prediction uncertainty estimation to achieve more reliable predictions, which can aid in identifying potential biases and increase trust and understanding of the deep model.

The performance of the proposed method might be further improved in several ways. In the current study, we utilized a variety of image modalities as model input to predict patient anatomy change, regardless of the available patient demographic, disease characteristics data (Table 1). Integrating both image data and patients' electronic health record data with a multimodal transformer could lead to a better solution. Recently, the advent of MR-guided radiotherapy in various institutions has significantly enhanced the visibility of soft tissue changes during treatment. The integration of MR data collected during radiotherapy could facilitate the development of a more precise and inclusive model for predicting anatomical changes, thereby aiding the decision-making process for ART.

## Conclusion

We constructed a transformer-based models (TransAnaNet) using planning dose map, initial treatment radiological image, and targets' structures from planning CT and initial CBCT to predict the anatomy change of treated region in later fractions. The proposed TransAnaNet has demonstrated promising capability in predicting the patient anatomy on later CBCT in the form of 3D image, which could be valuable to assist in the optimization of the workflow and the use of clinic resources related to ART. Future work is needed to incorporate OARs that are susceptible to notable anatomy change and of high radiosensitivity into our model, construct longitudinal prediction model, and validate the proposed methods in a prospective manner.

## Acknowledgements

We acknowledge the support from NIH (R01 CA251792).

## Conflict of Interest Statement

The authors have no conflicts of interest to declare that are relevant to the content of this article.

## Supporting Material

Please refer to Supplementary.docx for more details.

## Tables and Figures

**Table 1.** Patient and treatment characteristics.

**Table 2.** Image and structure similarity between CBCT21 to planning CT, CBCT01, and predicted CBCT21 from different anatomy change prediction models. (a) The image similarity is quantified with mean square error (MSE) and structure similarity index (SSIM). (b) The structure similarity is quantified with dice coefficient (dice) and average symmetric surface distance (ASD, unit: mm) for binary masks of patients' head and neck regions (body), primary tumors (GTVp) and involved lymph nodes (GTVn). Anatomy change prediction models built with Swin Transformer-based encoder (TransAnaNet), convolutional neural network based encoder (CNN), and basic vision-transformer-based encoder (ViT) are listed for comparison. Best performance for each metric is bolded.

**Table 3.** Performance comparison of using different baseline image and combination of image modalities as input for the proposed anatomy prediction model. a) The image similarity is quantified with mean square error (MSE) and structure similarity index (SSIM). b) The structure similarity is quantified with dice coefficient (dice) and average symmetric surface distance (ASD, unit: mm) for binary masks of patients' head and neck regions (body), primary tumors (GTVp) and involved lymph nodes (GTVn). Best performance for each metric is bolded.

**Figure 1.** The overall framework (a) and detailed architecture (b) of the proposed Transformer-based head and neck cancer patient anatomy change prediction (TransAnaNet) model. The hybrid Transformer-ConvNet network takes seven inputs: planning CT image, GTVp mask on planning CT, GTVn mask on planning CT, planned dose map, initial fraction CBCT (CBCT01), and GTVp and GTVn mask on it. The network predicts a nonlinear warping deformation vector field (DVF), which is then applied to CBCT01 image through a spatial transformation function to generate the predicted patient anatomy (fraction 21). For training data, the image similarity and structure similarity between the predicted and real CBCT21 are used as part of the loss functions to update the model, they are also used to evaluate the performance of the constructed model.

**Figure 2.** Qualitative performance evaluation of the proposed method for anatomy change prediction. Testing patients with a) the smallest, b) median, and c) the largest volumetric change between planning CT and CBCT21 are selected for comparison. CT, CBCT01, CBCT21 and predicted CBCT21 images are listed for comparison, original and predicted GTVp and GTVn are delineated in red and blue respectively. Difference images between the body masks of planning CT, CBCT01, and predicted CBCT21 to those from the real CBCT21 are shown in the second row for each patient.

**Figure 3.** Qualitative performance evaluation of the proposed method for anatomy change prediction with the planning CT image as the baseline image. Planning CT, CBCT01, CBCT21 and predicted CBCT21 from the proposed method for seven different testing patients (columns a~g) are listed as examples. The dash lines aside of each set of images show the width of the corresponding CBCT21 body contours, the red arrows point to the regions of obvious volumetric changes.

**Table S1.** Imaging protocols for planning CT, CBCT01, and CBCT21.

|  |  | Planning CT | CBCT01 | CBCT21 |
|---|---|---|---|---|
| **Machine** |  |  |  |  |
|  |  | Philip Brilliance Big Bore | Varian TrueBeam Ob-Board Imager |  |
| **Slice Thickness (mm)** |  |  |  |  |
|  | Median | 3.000 | 1.990 | 1.990 |
|  | IQR | 3.000 to 3.000 | 1.990 to 1.990 | 1.990 to 1.990 |
|  | Range | 1.500 to 3.000 | 1.989 to 1.992 | 1.990 to 1.992 |
| **Pixel Spacing (mm)** |  |  |  |  |
|  | Median | 1.171 | 0.511 | 0.511 |
|  | IQR | 1.171 to 1.171 | 0.511 to 0.511 | 0.511 to 0.511 |
|  | Range | 1.171 to 1.367 | 0.511 to 0.512 | 0.511 to 0.512 |
| **KVP** |  |  |  |  |
|  | Median | 120 | 100 | 100 |
|  | IQR | 120 to 120 | 100 to 100 | 100 to 100 |
|  | Range | 120 to 120 | 100 to 125 | 100 to 125 |
| **Exposure (mAs)** |  |  |  |  |
|  | Median | 300 | 150 | 150 |
|  | IQR | 300 to 300 | 150 to 150 | 150 to 150 |
|  | Range | 299 to 300 | 74 to 751 | 145 to 751 |

**Table S2.** Days between baseline CT scans (Planning CT), initial treatments (CBCT01), and treatment fraction 21.

|  | Planning CT to CBCT01 | CBCT1 to CBCT21 |
|---|---|---|
| Median | 12 | 28 |
| IQR | 11 to 14 | 28 to 29 |
| Range | 7 to 18 | 21 to 34 |

**Figure S1.** Sample images of planning CT, CBCT01, and CBCT21. Contours of primary tumor volumes and involved nodes are provided. Red lines delineate the primary tumors, blue lines delineate the involved nodes.

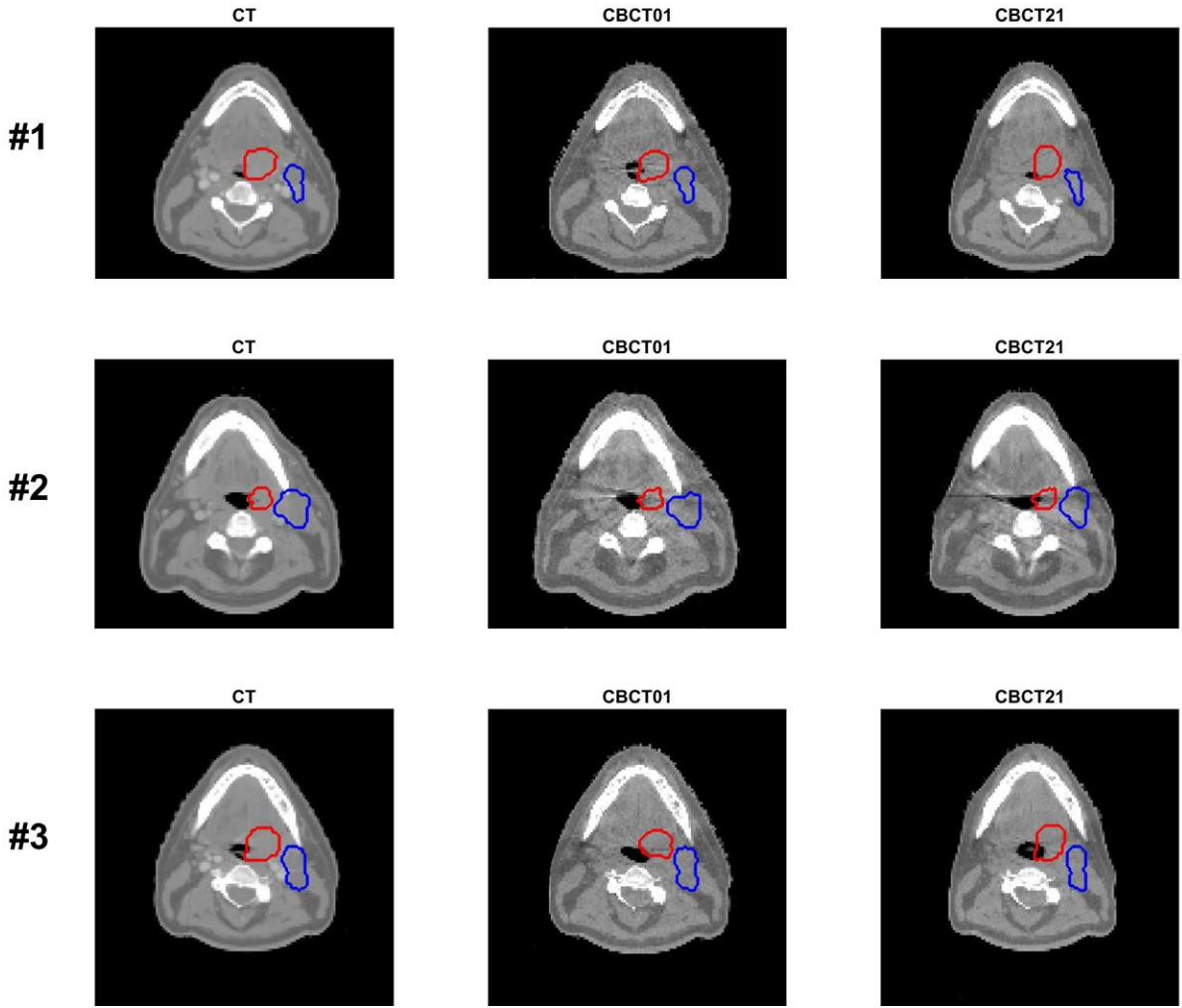

**Figure S2**. Distribution of patient BMI and BMI change. a) BMI before treatment, during treatment (fraction 21), and end of treatment. b) BMI net change between fraction 21 and treatment end to treatment start. c) BMI relative change between fraction 21 and treatment end to treatment start

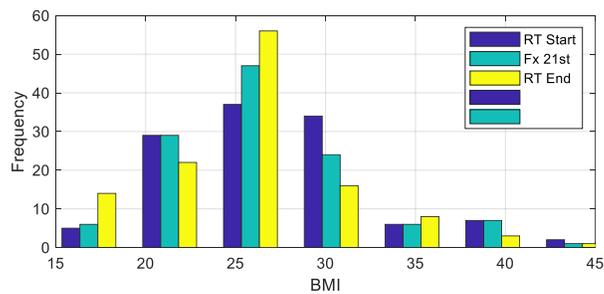

(a)

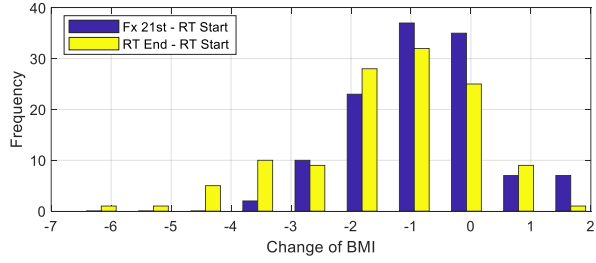

(b)

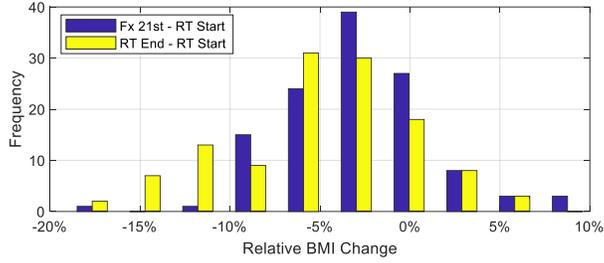

(c)

**Figure S3**. Distribution of model performance evaluation metrics on testing cohort. a) MSE for evaluate image similarity. b) SSIM for evaluation image similarity. c) Dice coefficient for evaluate patient body contour similarity. d) Dice coefficient for evaluate patient GTVp contour similarity. e) Dice coefficient for evaluate patient GTVn contour similarity.

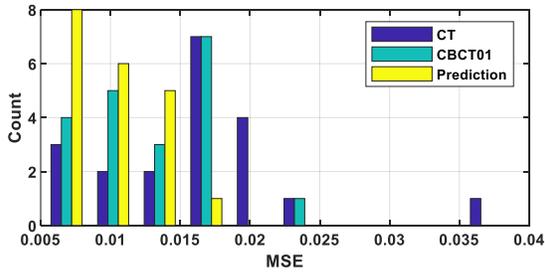
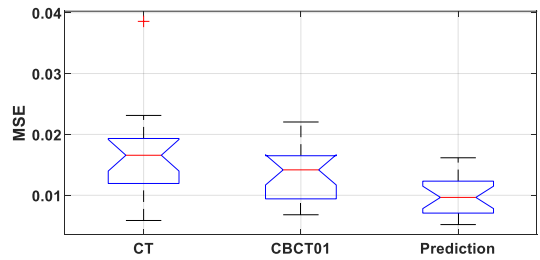

(a)

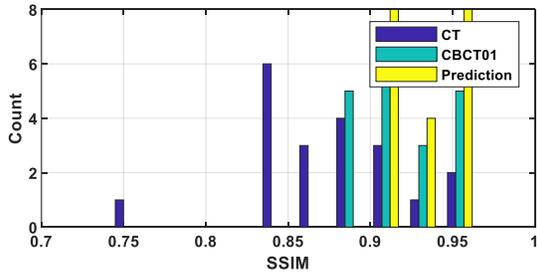
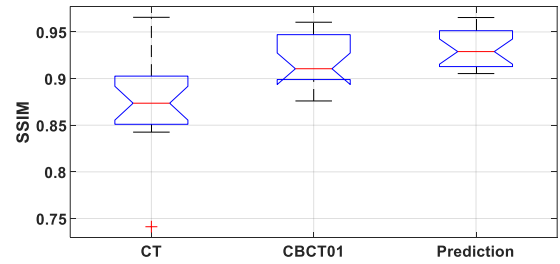

(b)

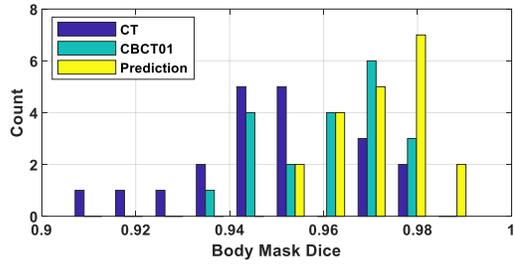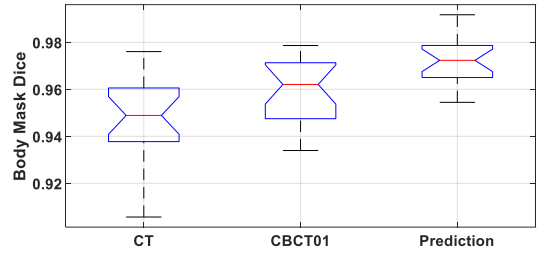

(c)

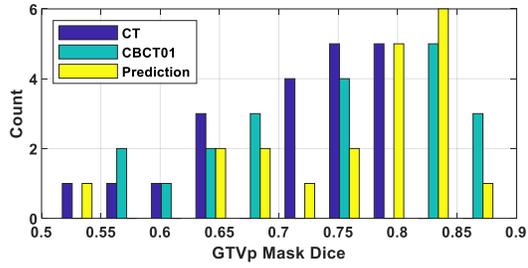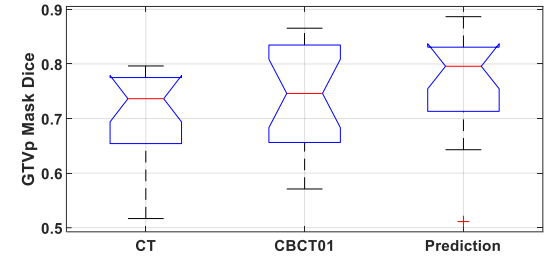

(d)

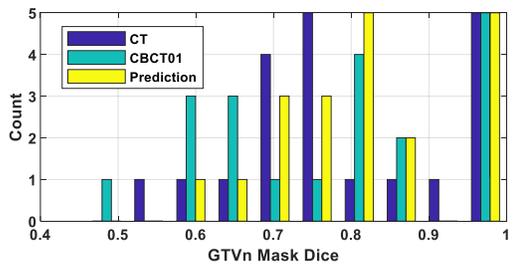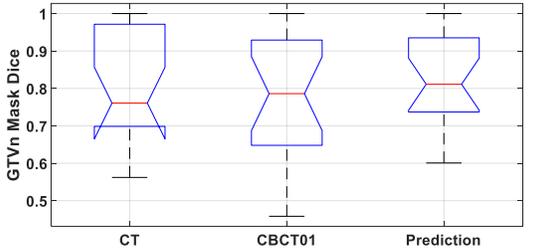

(e)